\documentclass[10pt,onecolumn,letterpaper]{article}

\title{Finding Islands of Predictability in Action Forecasting}

\newcommand{\argmax}{arg\,max}
\newcommand{\argmin}{arg\,min}
\usepackage{graphicx}
\usepackage{amsmath}
\usepackage{amssymb}
\usepackage[font=small]{caption}
\usepackage{amsmath}
\usepackage{enumerate}
\usepackage{makecell}
\usepackage{caption}
\usepackage[font=small]{subcaption}

\begin{document}
\author{
    Dan Scarafoni \\
    \texttt{danscarafoni@gatech.edu}
    \and
    Irfan Essa \\
    \texttt{irfan@gatech.edu}
    \and
    Thomas Pl{\"o}tz \\
    \texttt{thomas.ploetz@gatech.edu}
}

\maketitle

\setlength{\abovedisplayskip}{-1pt}
\setlength{\belowdisplayskip}{0pt}
\setlength{\abovedisplayshortskip}{-1pt}
\setlength{\belowdisplayshortskip}{0pt}

\setlength{\abovecaptionskip}{1ex}
\setlength{\belowcaptionskip}{1ex}
\setlength{\floatsep}{1ex}
\setlength{\textfloatsep}{1ex}

\begin{abstract}
We address dense action forecasting: the problem of predicting future action sequence over long durations based on partial observation.
Our key insight is that future action sequences are more accurately modeled with variable, rather than one, levels of abstraction, and that the optimal level of abstraction can be dynamically selected during the prediction process.
Our experiments show that most parts of future action sequences can be predicted confidently in fine detail only in small segments of future frames, which are effectively ``islands'' of high model prediction confidence in a ``sea'' of uncertainty.
We propose a combination Bayesian neural network and hierarchical convolutional segmentation model to both accurately predict future actions and optimally select abstraction levels.
We evaluate this approach on standard datasets against existing state-of-the-art systems and demonstrate that our ``islands of predictability'' approach maintains fine-grained action predictions while also making accurate abstract predictions where systems were previously unable to do so, and thus results in substantial, monotonic increases in accuracy.
\end{abstract}

%%%%%%%%% BODY TEXT
\vspace{-1em}
\section{Introduction}
\vspace{-1em}
Dense action forecasting aims to understand and anticipate future action sequences before they occur given an initial observation period. This is vital for a broad spectrum of problems. 
Robots collaborating with humans, for example, must be able to anticipate actions in order to safely work with their partners \cite{albrecht_autonomous_2018,heard_diagnostic_2018}.
Long term action sequences, in particular, are often very unpredictable, and predictors must adapt to the fundamental uncertainties of the video \cite{zhao_diverse_2020,ng_forecasting_2020,abu_farha_uncertainty-aware_2019,morais_learning_2020}.
One solution is to use coarser-grained labels to ``hedge a bet'' and predict an action abstraction when a concrete, fine-grained prediction is impossible. \cite{suris_learning_2021}.
\begin{figure*}[t]
	\centering
	\begin{subfigure}[t]{.49\textwidth}
	\centering
	    \includegraphics[width=.75\textwidth, height=4cm]{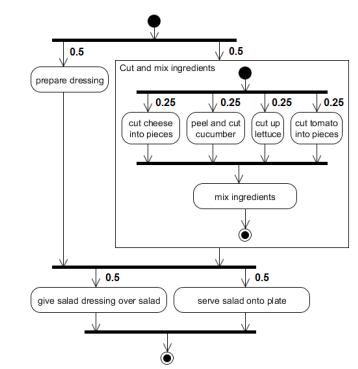}
	    \subcaption{The activity diagram from which action sequences were sampled for 50 Salads demonstrates that many action sequences are, to a degree, random by design \cite{stein_combining_2013}}
	    \label{fig:titlea}
	\end{subfigure}
	\begin{subfigure}[t]{.49\textwidth}
	    \includegraphics[width=1\textwidth, height=3cm]{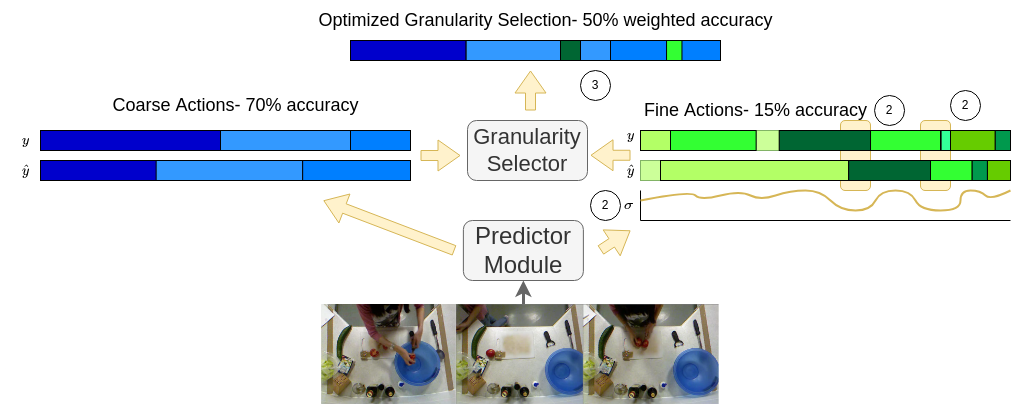}
    	\subcaption{We can use uncertainty estimation (1) to find the predictable areas (2) and predict correct coarse-grained actions elsewhere to improve forecasting accuracy (3).}
		\label{fig:titleb}
	\end{subfigure}
	\vspace{-1em}
\end{figure*}

We propose a system to find and replace inaccurate, fine-grained future action sequence predictions with accurate coarse-grained ones.
This system extends uncertainty estimation to dense action forecasting and leverages temporal convolutional networks to select the correct level of abstraction at each time step to maintain high prediction accuracy.
Our experiments show that most parts of future action sequences can be predicted confidently in fine detail only in small segments of future frames, which are effective ``islands'' of high model prediction confidence in a ``sea'' of uncertainty.
This we contrast with existing methods which try to make fine-grained predictions at every point in the future \cite{Farha2018b,zhao_diverse_2020}.
In the standard 50 Salads dataset, for example, ingredients can be chopped in many different orders; however, dressing is almost always poured on the salad at the end.
Thus, predicting what ingredients will be chopped next is often impossible (see Fig. \ref{fig:titlea}), but it is almost certain that the dressing will be poured onto the salad near the end of the video.

Our system gives ``partial credit'' for guesses where correct coarse actions are substituted for fine.
Since the relative value of coarse or fine predictions varies depending on the specific use case, and so, there is a need to bias the system towards coarse or fine-grained guesses depending on which preference will ultimately maximize performance.
To this end, we introduce Granularity Loss, which further tunes predictions to the data and evaluation scheme.
We perform extensive testing across all possible evaluation schemes with standard metrics on benchmark datasets and demonstrate that our method substantially improves performance on state-of-the-art systems.
Fig. \ref{fig:titleb} demonstrates that substantial gains in accuracy can be made using our system.\\
Our primary contributions are (1) a novel system that leverages uncertainty estimation and hierarchical temporal segmentation to find the predictable sections (``islands'') of future action sequences and substitute accurate, coarse-grained predictions where none such exist.
(2) a novel loss function, Granularity Loss, for tuning the system to specific evaluation schemes.
(3) The introduction of uncertainty estimation to the dense action forecasting problem domain.
(4) Extensive experimentation on standard datasets demonstrates that our method substantially improves dense action forecasting for state-of-the-art systems.

\vspace{-1em}
\section{Related work}
\vspace{-1em}
Future action prediction has seen explosive growth in recent years.
Our work deals with dense action forecasting: the prediction of long-term (on the order of several minutes), action sequences and their durations \cite{sener_temporal_2020,zhao_diverse_2020,piergiovanni_adversarial_2020,sener_zero-shot_2019,guan_generative_2020}.

Despite significant gains, dense action forecasting remains very difficult, rarely exceeding $30\%$ accuracy for the longest term predictions on standard data sets \cite{ng_forecasting_2020,gammulle_forecasting_2019,piergiovanni_adversarial_2020}.
Some datasets are known to be highly unpredictable and were designed as such (see Fig. \ref{fig:titlea}) \cite{stein_combining_2013,morais_learning_2020}.
Recent work has shown that often coarser-grained actions can be reliably predicted even when fine-grained actions cannot \cite{suris_learning_2021}.
This motivates accurate prediction of coarser-grained activity labels outside of the ``islands.''
In this work, we posit a method to estimate such predictable ``islands'' and make multiple-granularity predictions to accurately estimate the future.

\vspace{-1em}
\subsection{Uncertainty estimation in video action prediction}
\vspace{-1em}
Predicting the correctness of forecast actions requires uncertainty estimation.
More generally, uncertainty estimation is an important area of research in computer vision.
Equipping a model with confidence about its prediction allows a system to understand and cope with sensor and model error.
Classification logits do not necessarily reflect model confidence, and thus more sophisticated methods are necessary \cite{gal_dropout_2016}.
Generalized methods for uncertainty estimation in computer vision have made great strides in recent years, and there have been many attempts to model dense action forecasting systems with probability distributions \cite{zhao_diverse_2020,piergiovanni_adversarial_2020,mehrasa_variational_2019,Farha2019b}; However, no such techniques exist for dense action forecasting \cite{gal_dropout_2016,lakshminarayanan_simple_2017}.
There is a need to develop uncertainty estimation for this problem, as prediction confidence is essential to ``hedging bets'' as to how precise a prediction should be \cite{suris_learning_2021}.

\vspace{-1em}
\section{Finding islands of predictability}
\vspace{-1em}
\subsection{Problem statement}
\vspace{-1em}
Given per-frame action predictions for multiple action granularities, our system selects the ones that both maximize accuracy and granularity.
Formally, given a video of $T$ frames with $t_0$ observed frames, $g \in G$ levels of granularity, and a prediction for each granularity level in data point $n$ at frame $t$ $\hat{a}^{t,n}_g$, the goal is to design a Granularity Selector system $H$ with parameters $\theta$ which maximizes
\begin{align}
\theta^* = \argmax_{\theta} \sum_{n=1}^N\sum_{t=t_0^n+1}^T \mathcal{A}(H(\hat{a}^{t,n}_1 \dots \hat{a}^{t,n}_G; \theta), a_{g^*}^{t,n}, \beta) \end{align}
$\mathcal{A}$ is a weighted accuracy metric \cite{suris_learning_2021}.
It is parameterized by $\beta$, which determines the accuracy weight or ``partial credit'' given to correct coarser levels of granularity if the fine-grained actions are not predicted.
$H$ is the Granularity Selector which selects a final granularity for each frame from options at different levels
\begin{align}
    H: \hat{a}_1 \dots \hat{a}_G \rightarrow \hat{a}_{g^*}
\end{align}

\subsection{Discovering islands of predictability}

\begin{figure*}[t]
    \centering
    \includegraphics[width=.99\textwidth, height=3cm]{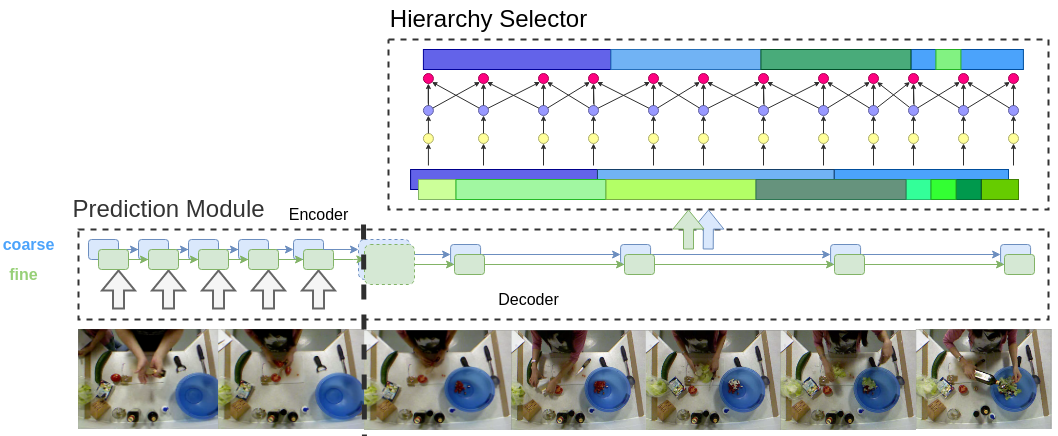}
    \vspace{1em}
    \caption{An overview of our approach. Video features serve as input to the Prediction Module, which makes future action predictions across multiple levels of granularity. The probabilities for each action serve as input to the Granularity Selector, which ultimately decides at what level of abstraction each frame will be labeled.}
    \label{fig:bnn}
\end{figure*}

We present a two-part system for finding islands of predictability, where we first a Prediction Module predicts action labels and durations across different granularity levels, and then a Granularity Selector chooses final granularities from these.
We show an overview of our system in Fig. \ref{fig:bnn}.

The Prediction Module is a dense action forecaster which predicts future actions over long periods (on the order of several minutes).
Formally, given a series of $t_0$ observed frames $x_1\dots x_{t_0}$, which correspond to a list of activities $a_1\dots a_n$, we want to predict the remaining actions $a_{n+1}\dots a_N$ as well as the time lengths for these actions $\ell_{n+1}\dots \ell_N$.
The predictions then serve as input to the Granularity Selector.

% The Granularity Selector chooses the finest-grained and most likely to be correct action for a given time step.
% Given the choice between a coarse and fine-grained action, the system should choose the fine-grained if it is likely to be correct, and the coarse otherwise.
% This motivates the need for estimating prediction confidence, which is uncertainty estimation.

Any classifier can have its outputs converted into confidence measurements with ensembling \cite{gal_dropout_2016,lakshminarayanan_simple_2017}.
We improve upon this notion by designing a novel Bayesian neural network Prediction Module that outputs confidence scores for each prediction.
This represents the output layer $f(x)$ as a probability distribution $f(\textbf{x}; \theta) \equiv p(\cdot, x)$.
The action output for each class in each step is a Gaussian distribution representing a probabilistic distribution for each class.
\begin{align}
\hat{a}^{t,n}_c \sim N(\mu_{t,c}, \sigma_{t,c}) ~ t \in T, c \in C, n \in N
\end{align}
where $\hat{a}^{t,n}_c$ is the logit value of class $c$ at time step $t$ for data point $n$.
This allows the logits to be treated as true measures of prediction certainty.
Our RNN also must take into account the fact that the probability distribution at prediction $m$ becomes not only a function of the input data, but previous predictions as well
\begin{align}
f(\textbf{x}, \textbf{m}; \theta) \equiv p(\cdot, x, \hat{a}^{1,n}, \dots \hat{a}^{m-1,n})
\end{align}
After replacing the output classifier with the Bayesian modification, it can be trained as a BNN with a simple probabilistic modification to the loss function \cite{lakshminarayanan_simple_2017}.

This network is trained not on the outputs of each action step, altered to be a Dirichlet distribution $Dir(y|\alpha)$ with concentration parameters $\alpha j > 0$, a categorical distribution
\cite{gast_lightweight_2018}.
Our network is the first of its kind to allow for confidence measurements of long-term action predictions.
Not only is this essential for selecting what level of granularity to use, but, more fundamentally, it allows for informed estimations of the predictability of the future.

The Granularity Selector then selects the final granularity.
It is most important to select fine-grained predictions when the system has high confidence in its guesses and coarse otherwise.
The input of this network is the predicted mean probabilities for each fine-grained class $c_f \in C_f$ and each coarse-grained class $c_c \in C_c$ at each prediction step $o_t$ for $t \in [t_0+1 \dots T]$:
\begin{align}
o_t = [ \mu_{1_f} \dots \mu_{C_f} \mu_{1_c} \dots \mu_{C_c}]
\end{align}
Picking hierarchy labels should  be as follows:
A system should select the fine-grained label whenever correct, and the coarse label otherwise.
Formally, we choose labels $y$ as such:
\begin{align}
     y^{t,n} = \begin{cases}
    0 & \text{if } \hat{a}_1^{t,n} = a^{t,n}_1\\
    1              & \text{otherwise}
\end{cases}
\label{alg:labelchoice}
\end{align}
Where $a^{tn}_1$ is the ground truth fine-grained label.
This allows labels to be derived quickly and intuitively from any existing dataset and Prediction Module.
This corresponds to action segmentation in that it takes, as input, video frame features and labels each with an action.
Any feed-forward classifier can be used for the Granularity Selector, but the temporal context has shown itself to be invaluable in video segmentation \cite{li_ms-tcn_2020,Chang2019}.
We, therefore, use a temporal convolutional segmentation network to gather features over both temporal and hierarchical dimensions to make a more refined Granularity Selector. 
To train the network we use standard cross entropy loss:
\begin{align}
\mathcal{L}_{A}(\hat{y}, y) = \frac{1}{N}\sum_{n=1}^{N}\frac{1}{T_n}\sum_{t=1}^{T_n} -log(\hat{y}_{g^*}^{n,t})
\end{align}
Where $\hat{y}_{g^*}^{n,t}$ is the probability for the true granularity.
Ideally, all correct guesses should have high confidence, but, in practice, the correlation between confidence and accuracy is not perfect.
There is, therefore, a need to bias the system towards coarse guesses or fine guesses depending on their relative value when such decisions are unclear.
To this end, we introduce the novel Granularity Loss, $\mathcal{L}_{G}$, which penalizes coarse action selection.
\begin{align}
\mathcal{L}_{G}(\hat{y}, y) =\frac{1}{N}\sum_{n=1}^{N}\frac{1}{T_n}\sum_{t=1}^{T_n} -log(\hat{y}_{g^*}^{t,n})I(\hat{y}^{t,n}= 0)
\end{align}
Where $I$ is the identity function mapping to $1$ when the statement within is true and 0 otherwise, and $y^{n,t}$ is the true label.
$\hat{H}$ is a variation of $H$ that takes the probability features $O$ and outputs the correct granularity level

This brings the total optimization function to 
\begin{align}
\theta^* = \argmin_{\theta} \mathcal{L}_{A}(\hat{H}(O; \theta), y) + \gamma\mathcal{L}_{G}(\hat{H}(O; \theta), y)
\end{align}
where $\gamma$ is the granularity penalty i.e. the additional weight placed on the fine-grained actions versus coarse.

\vspace{-1em}
\section{Experiments and results}
\vspace{-1em}
In this section, we will demonstrate the performance gains of this methodology across multiple action Prediction Modules and Granularity Selector versus state-of-the-art dense action forecasting.

\vspace{-1em}
\subsection{Datasets and metrics}
\vspace{-1em}
We evaluate our model on Breakfast and 50 Salads, two leading benchmarks for action prediction \cite{Kuehne2014a,stein_combining_2013}.
The Breakfast dataset contains 712 videos, each several minutes in length and shot with multiple camera angles. 
Each video belongs to one out of ten activities, such as making coffee or scrambling eggs.
The frames are annotated with fine-grained labels like ``pour cream'' and ``grab cup''. 
There are 48 actions total.
For evaluation, we use the standard 4 splits as proposed.
%, each with 252 videos used for tests and the rest used for training.
% As such, these two datasets fit the domain area optimally.

The 50 Salads dataset contains 50 videos showing people preparing salads. 
The average duration is 6.4 minutes, and 20 actions per video.
In addition, to start and end action labels, there are 17 fine-grained action labels such as ``cut cucumber'' and ``add oil.'' 
For evaluation, we use five-fold cross-validation and report the average where ten videos are held out for a test in each split.
We design coarse label sets for 50 Salads and Breakfast of five and three labels each.
These action sets are designed to be highly predictable and non-trivial.
By finding and substituting inaccurate fine-grained labels for these accurate, coarse-grained labels, we demonstrate that our methodology allows for monotonic, often substantial, gains in accuracy, regardless of the value of substituted coarse labels versus fine.

% These are labels of five and three actions for 50 Salads and breakfast, respectively.
% The labeling details are shown in Fig. \ref{tab:dataset}.

% \begin{table*}[b]
%     \centering
%     \begin{tabular}{c|p{5cm}|p{5cm}}
%       Dataset & Fine actions & Coarse actions \\\hline
%       50 Salads  & \parbox{5cm}{peel cucumber, cut cucumber, place cucumber into bowl, cut tomato, place tomato into bowl, cut cheese, place cheese into bowl, cut lettuce, place lettuce into bowl, mix ingredients, add oil, add vinegar, add salt, add pepper, mix dressing, serve salad onto plate, add dressing, action start, action end} & cut and mix ingredients, prepare dressing, serve salad, action start, action end\\
%       Breakfast & 48 & 3 \\ 
%     \end{tabular}
%     \caption{Label statistics for each data set}
%     \label{tab:dataset}
% \end{table*}

The standard metric for dense action forecasting is mean over class accuracy \cite{Farha2018b}. 
When coarse granularity is allowed for substitute predictions, it is generally interpreted as ``hierarchical'' action recognition or prediction, where coarser granularity measures are allowed to count as ``partial credit'' instead of finer-grained guesses \cite{suris_learning_2021,long_searching_2020}.
For our experiments, we parameterize the weighted accuracy with $\beta$ which represents the weight of coarser-grained predictions.
As with previous work, we experiment with valuing coarse less than fine grained ($\beta < 1$), equal to fine grained ($\beta = 1$), and more than fine grained ($\beta > 1$) \cite{suris_learning_2021,wu_hierarchical_2019}.

\vspace{-1em}
\subsection{Finding islands of predictability for dense action forecasting}
\vspace{-1em}
Tab \ref{tab:top1} shows state of the art dense forecasting performance.
Without ground truth observation labels at test time, few exceed $20\%$ accuracy, and none can exceed $40\%$ accuracy.
This underscores the difficulty of this task.
We note that with access to ground truth observations at test time, prediction accuracy does increase \cite{gammulle_forecasting_2019,Farha2018b}.
However, in real scenarios, these labels will not be present, and so we argue that it is of paramount importance to focus on non-ground-truth based systems.
Accordingly, we emphasize non-ground-truth-based experiments in this work.

\begin{figure*}[h]
	\centering
    \includegraphics[width=1\textwidth]{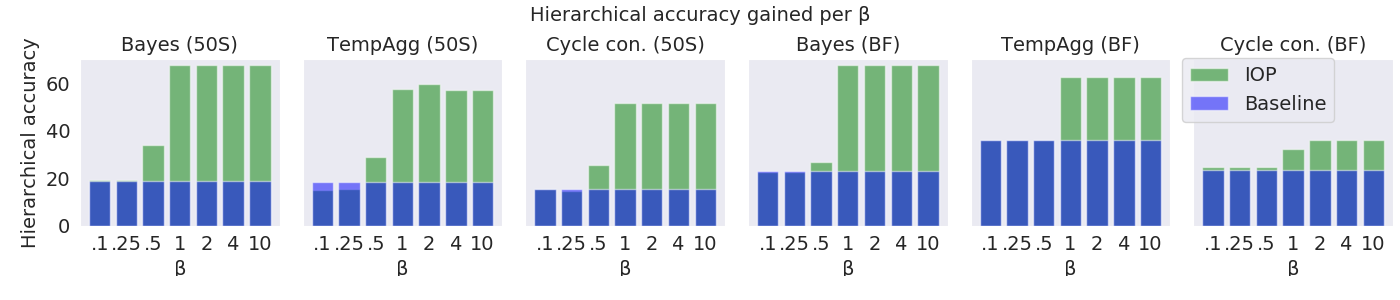}
    \caption{Accuracy improvements via islands of predictability (green) over state-of-the-art action forecasting (blue). Results were evaluated on 50 Salads (left) and Breakfast (right). Results show that hierarchical selection provides additional gains in accuracy across multiple baselines, with gains and overall performance being greatest for the proposed (Bayes) system and on the 50 Salads dataset.}
    \label{fig:main_results}
\end{figure*}

We demonstrate that our system can identify these inaccurate segments ahead of time and substitute far more accurate coarse-grained labels in their place to improve performance.
At the same time, we show that our system does not reduce performance.
Thus, we show that finding the islands of predictability leads to substantial monotonic gains in performance.
We show our results in Fig. \ref{fig:main_results}.
For this experiment, we evaluate Granularity Selectors across all values $\beta$ and demonstrate performance over all possible preferences.
We demonstrate the efficacy of three action prediction systems.
\textbf{Bayes} is our BNN predictor.
\textbf{Cycle Consistency} is the action predictor from \cite{farha_long-term_2020}. 
This system uses segmentation to extract features and a GRU with cycle consistency to make future action predictions.
\textbf{Temporal Aggregation} implements a multi-granular encoder to make predictions \cite{sener_temporal_2020}. 
Unlike the other baselines, we implement the version that takes ground-truth labels as input to demonstrate the effectiveness of our framework on ground-truth input as well as non-ground-truth. 

For both baselines, prediction uncertainty is estimated via the ensembling of three separately trained classifiers.
In the graphs, the blue bar shows the baseline system accuracy, while the green bar represents additional accuracy gained from the Granularity Selector.
Note that as $\beta$ approaches zero, accuracy approaches the fine-grained accuracy. 
Similarly, as $\beta$ approaches infinity, it approaches the coarse-grained accuracy.
Results demonstrate that finding islands of predictability adds increased accuracy, often substantially.
This is especially true for benchmark values of $\beta$ (0.5, 1, and 2) \cite{suris_learning_2021}.
Across four of six systems, our proposed system doubles at least one accuracy value.
The two exceptions, Temporal Aggregation- Breakfast and Cycle consistency- Breakfast, still have double-digit improved accuracy scores.
The proposed Bayesian network achieves the largest increase in accuracy, the highest overall accuracy for both datasets, as well as top accuracy for $\beta=0.5$ on 50 Salads.

In only one experiment, Temporal Aggregation- 50 Salads, does the Granularity Selector reduce performance slightly.
We notice that 50 Salads has a higher relative improvement from the Granularity Selector, which is to be expected since its fine-grained labels are more unpredictable and can thus benefit more from coarse-grained labels \cite{morais_learning_2020}.

\begin{table*}
    \centering
    % \begin{subtable}[t]{0.95\textwidth}
    \begin{tabular}{c|c|c|c|c}
        & \multicolumn{2}{c}{50 Salads} & \multicolumn{2}{c}{Breakfast} \\\hline
        \textbf{System} &\textbf{20\%/50\%} & \textbf{30\%/50\%} & \textbf{20\%/50\%} & \textbf{30\%/50\%} \\\hline
        RNN \cite{Farha2018b} & 13.49 & 9.77 & 15.82 & 19.21\\
        CNN \cite{Farha2018b} & 9.87 & 10.86 & 14.54 & 18.76 \\
        Uncertainty aware \cite{abu_farha_uncertainty-aware_2019} & 12.82 & 12.31 & 14.20 & 16.86 \\
        Time conditioned \cite{ke_time-conditioned_2019} & 15.99 & 15.59 & 15.84 & 19.75\\
        Cycle consistency \cite{farha_long-term_2020} & 15.25 & 15.89 & 21.54 & 25.20\\
        Adversarial grammar \cite{piergiovanni_adversarial_2020} & 21.2 & 19.8 & N/A & N/A\\
        Temporal Aggregation \cite{sener_temporal_2020} & 15.57 & 15.5 & 20.85 & 23.61 \\
        \textbf{Bayesian predictor (ours)} & 19.71 & 18.69 & 19.21 & 25.1 \\
        Attn-GRU \cite{ng_forecasting_2020} & \textbf{23.88} & \textbf{26.39} & 20.85 &
        23.61\\
        MAVAP \cite{loh_long-term_2022} & 16.1 & 14.8 & \textbf{32.9} & \textbf{35.5} \\\hline
    \end{tabular}
    \vspace{1em}
    \caption{MoC accuracy in previous works on the 50 Salads and Breakfast datasets are evaluated with standard protocols of observing $20-30\%$ of a video and predicting $50\%$ of the remaining video. Results never exceed 30\%, a fact which highlights the difficulty of this problem and the inherent unpredictability of the future.}
    \label{tab:top1}
\end{table*}

\vspace{-1em}
\subsection{Optimizing hierarchy selection}
% \vspace{-1em}

\begin{figure*}[bh]
	\centering
	\includegraphics[width=0.99\textwidth]{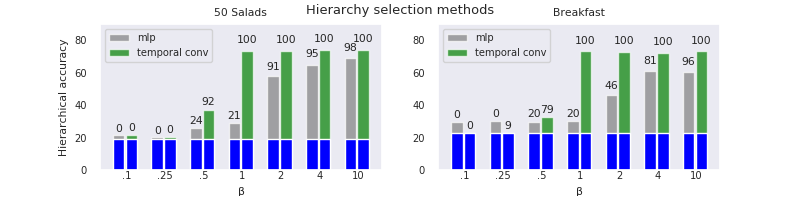}
	\vspace{0.5em}
	\caption{Hierarchy selection results on  50 Salads and Breakfast. Percent frames guessed coarse are displayed above each bar. 
	%Results show that The temporal convolutional network performs best, but that MLP can make more nuanced granularity decisions.
	}
	\label{fig:50shs}
\end{figure*}

We run experiments to optimize the Granularity Selector given a pretrained action Prediction Module.
As a baseline here, we consider a simple multilayer perceptron model and contrast it with the temporal convolutional segmentation model.
The results are shown in Fig. \ref{fig:50shs}.
Results indicate that the temporal convolutional network achieves superior performance, which indicates, by extension, that temporal context is important for hierarchy selection.
The temporal convolutional network tends to heavily favor either coarse or fine-grained classes.
The MLP tends to guess granularity more proportionally with $\gamma$, however, it is still less accurate.

\begin{figure*}[t]
	\centering
	\begin{subfigure}[t]{0.99\textwidth}
        \includegraphics[width=1\textwidth, height=3cm]{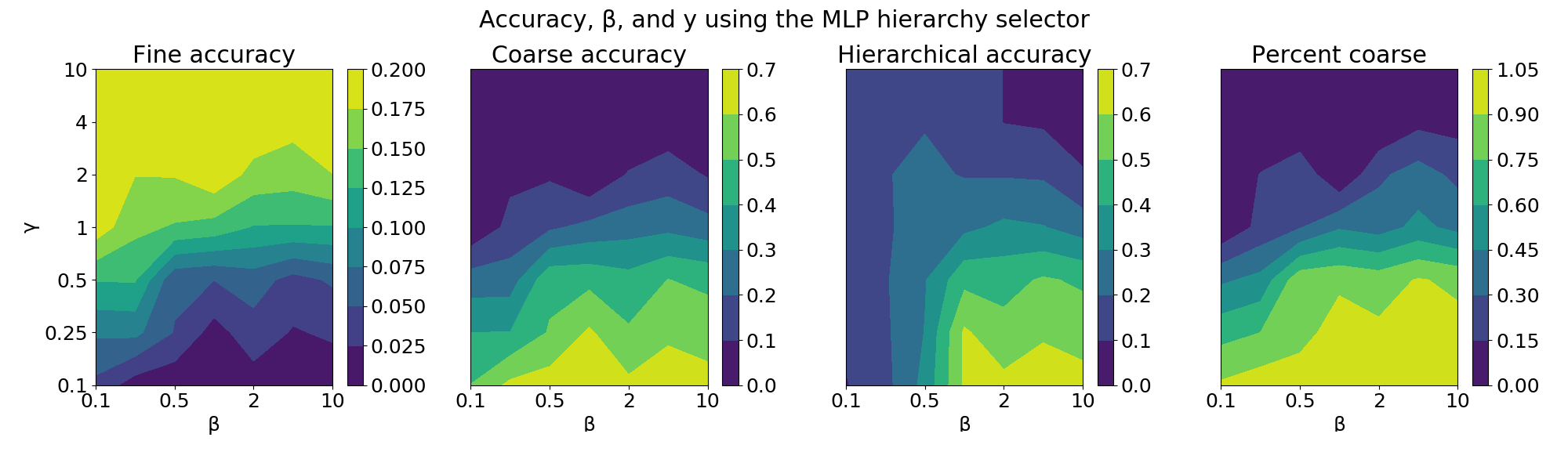}
    \end{subfigure}
    
    \vspace{-.75em}
    \begin{subfigure}[t]{.99\textwidth}
        \includegraphics[width=1\textwidth, height=3cm]{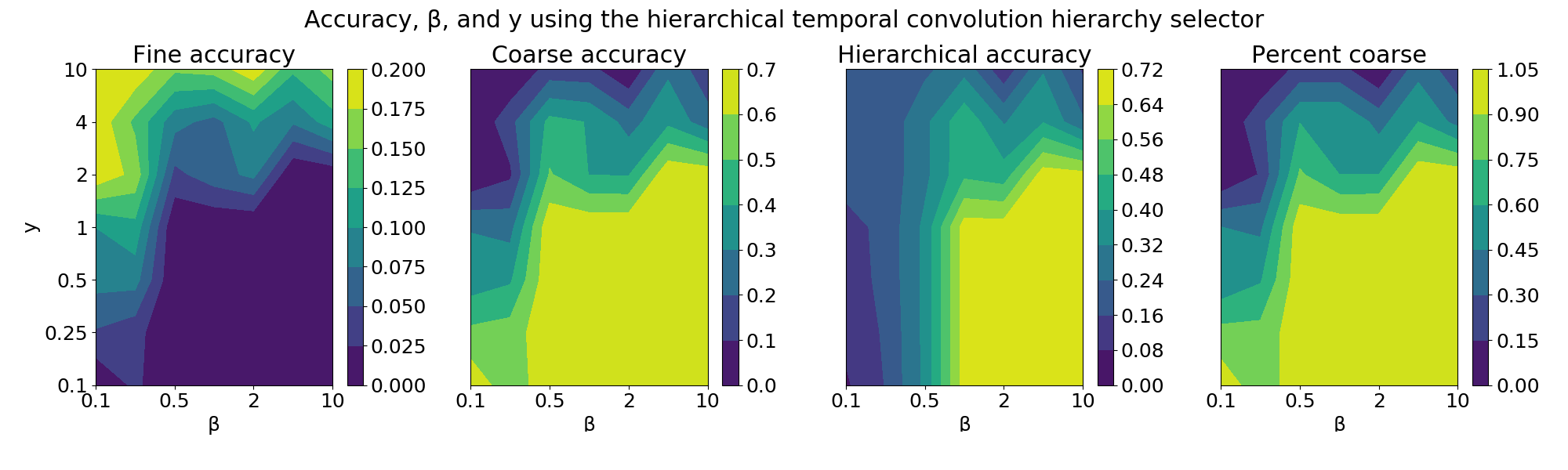}
    \end{subfigure}
    
    \caption{Heat maps demonstrating the relationship between granularity penalty ($\gamma$), $\beta$, and accuracy score on 50 Salads validation data. Increased $\gamma$ results in a decreased preference for coarse actions, and increased fine-grained accuracy, and shows that tuned Granularity Loss is necessary to optimize hierarchy selection.}
    \label{fig:gp_w_sweep}
\end{figure*}

\vspace{1em}
\subsection{The impact of Granularity Loss}
\vspace{-.5em}
To train an effective system, we must understand the relationship between the performance of that system and Granularity Loss.
Fig. \ref{fig:gp_w_sweep} shows how accuracy measurements (coarse, fine, and hierarchical) change with $\beta$ and $\gamma$.
Values for fine-grained accuracy peak when $\gamma$ is maximized with little influence from $\beta$, and coarse-grained accuracy similarly peaks when $\gamma$ is minimized.
Despite only being an evaluation metric, $\beta$ has a non-zero impact on weighted accuracy as validation weighted accuracy is used as a stopping criterion during training.
Results indicate intuitively that accuracy is maximized when the network is trained to have $\gamma$ proportional to the inverse of the weighted accuracy's $\beta$ value.
In all experiments, this is the value of $\gamma$ used unless otherwise stated.

\vspace{-1em}
\subsection{Qualitative results}
\begin{figure*}
    \vspace{-1em}
	\begin{subfigure}[b]{.99\textwidth}
	    \includegraphics[width=1\textwidth]{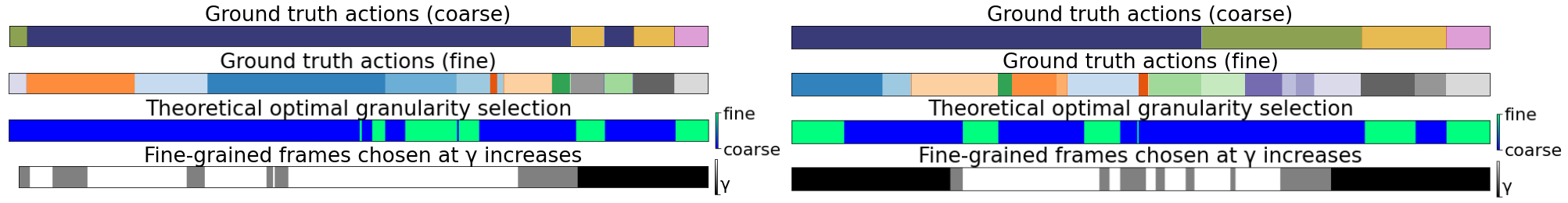}
	        \vspace{-2em}
	    \subcaption{50 Salads}
	\end{subfigure}
	\begin{subfigure}[b]{.99\textwidth}
	    \includegraphics[width=1\textwidth]{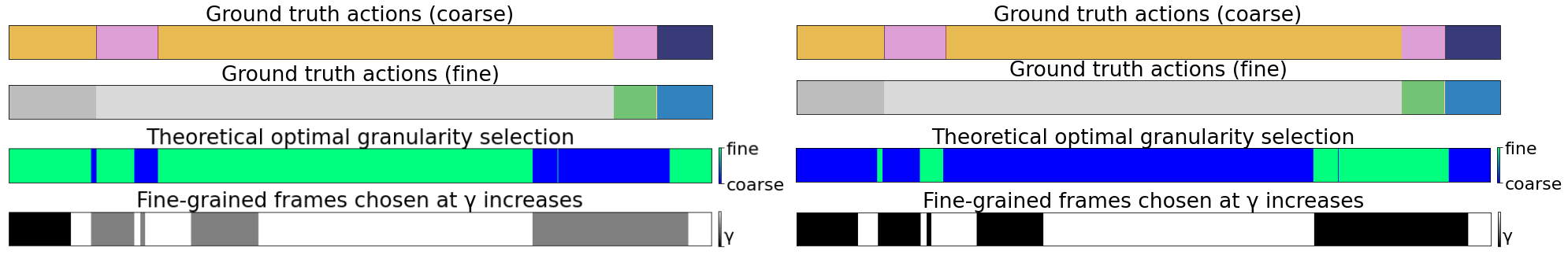}
	    \vspace{-2em}
	    \subcaption{Breakfast}
	\end{subfigure}
    \vspace{1em}
    
	\caption{Qualitative Results demonstrating ground truth fine-grained actions (top), best possible distribution of coarse and fine labels (middle), and hierarchies selected per frame as $\gamma$ increases (bottom). Results demonstrate not only evidence for the existence of ``Islands of Predictability'' in future action sequences, but also that increasing granularity penalty increases the bias towards classifying frames as fine-grained.}
	\label{fig:qualitative}
\end{figure*}

\vspace{-1em}
We analyze qualitative results in Fig. \ref{fig:qualitative}.
Each shows the ground truth actions (top), the best possible selection of coarse and fine labels as per Alg. \ref{alg:labelchoice} (middle), and hierarchies selected per frame over each value $\gamma$ (bottom).
Most of the optimal actions predicted for each frame are coarse-grained, with ``islands'' of predictability where fine-grained actions are the best choice.
This trend is not absolute, however, as the second Breakfast example shows an instance where most of the frames are classified as fine-grained.
Considering that this example is dominated by one long fine-grained action, it is the result of this action being correctly guessed by the dense action forecaster.

The bottom bar for each example shows which frames the Granularity Selector labels as ``fine-grained'' as $\gamma$ increases; the shade of color determines at what value $\gamma$ the selector begins selecting the fine-grained label for the frame.
We see that as $\gamma$ increases, the selector classifies more frames as ``fine-grained.''
In other words, the ``islands'' expand until they occupy the entirety of the predicted sequence.
This behavior is similar to islands in a shallow sea becoming ``larger and larger'' as water levels recede.
This is important behavior, as the Granularity Loss is designed to encourage the system to favor fine-grained predictions over coarse at higher values of $\gamma$.
% \vspace{-2em}

\vspace{-1em}
\subsection{Prediction confidence for dense action forecasting}
\vspace{-.5em}
For uncertainty estimation, we evaluate three novel uncertainty estimation systems: ensemble, dropout ensemble, and Bayesian.
The ensemble method trains an ensemble of classifiers, dropout repeatedly sampling from logits in a network with test-time dropout \cite{gal_dropout_2016}.

\begin{table}[h]
    \vspace{-1em}
    \centering
    \begin{tabular}{c|c|c}
        Method & NLL & NLL MSE \\\hline
        Ensemble & 0.040 & \textbf{0.0009} \\
        Dropout & 0.033 & 0.0023 \\
        Bayesian Network & \textbf{0.004} & 0.0046 \\
    \end{tabular}
    \vspace{-0.5em}
    \caption{Prediction NLL for each forecasting network, as well as the difference in NLL to ground truth. 
    The Bayesian network performs best at minimizing uncertainty in prediction, but the Ensemble provides variance most similar to the ground-truth data.
    }
    \vspace{-.5em}
    \label{tab:nll_results}
\end{table}

\begin{figure}[h]
    \centering
    \includegraphics[width=.95\textwidth, height=3cm]{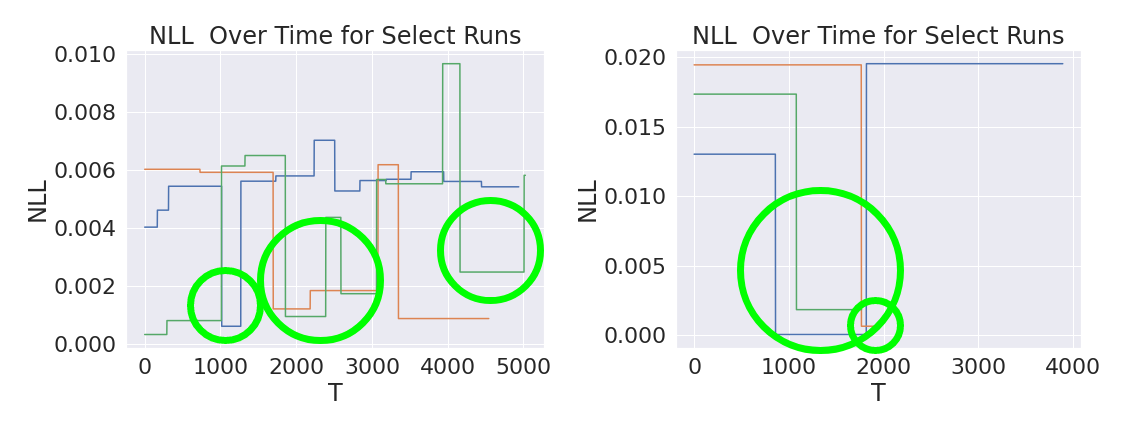}
    \caption{The NLL over time of select fine-grained 50 Salads (left) and Breakfast (right) sequences. ''Islands'' of predictability are highlighted.}
    \label{fig:nllq}
\end{figure}

We utilize the long-term action prediction protocols from previous works \cite{farha_long-term_2020} but use NLL (negative log-likelihood) instead of MoC accuracy as the metric. 
NLL is the standard metric for predictor uncertainty, and the smallest predicted value is best \cite{gal_dropout_2016,lakshminarayanan_simple_2017,gast_lightweight_2018}.
Measurements were taken from the 20\% observed and 50\% predicted experiment to best gauge effectiveness over long durations.
However, because of the inherent unpredictability of the future, there is arguably a need to also make diverse predictions, and thus not minimize NLL \cite{zhao_diverse_2020}.
We also, therefore, minimize the difference between the predicted NLL and the per-frame NLL of the entire dataset.
This first involves using linear interpolation to increase the length of all predicted and ground-truth actions to the length of the longest video.
This ensures that all videos can be compared despite differing lengths.
Then the NLL is calculated for each frame for both the ground truth and predicted sequences.
After this, we take the mean squared error between probability distributions of predictions ($\hat{p_a}$) and ground truth ($p_a$) over time:
\begin{align}
MSE_{NLL} = \frac{1}{N}\sum_{n=1}^N \frac{1}{T_n} \sum_{t=1}^{T_n} (\hat{p_a}^{n,t} - p_a^{n,t})^2
\end{align}
The results for 50 Salads are shown in Tab.\ \ref{tab:nll_results}.
Results indicate that our novel Bayes network performs the best in terms of minimizing NLL, but that the ensemble network performs best in terms of minimizing $MSE_{NLL}$.
This means that while the Bayes network optimizes certainty, the ensemble network best reflects the uncertainty of the ground truth data.
Fig. \ref{fig:nllq} shows the qualitative results of NLL over time, and shows evidence for the existence of islands of predictability with drastic dips in predicted NLL.
% \vspace{-60pt}

% \vspace{-400pt}
\section{Conclusion}                                                                            
By estimating the uncertainty of future predictions, we show that future actions are largely difficult to foresee in fine detail, with ``islands of predictability'' punctuating this sea.
By finding these islands and substituting coarse labels in other places, we substantially improve accuracy versus state-of-the-art systems.

\bibliographystyle{unsrt}
\bibliography{refs}

\begin{thebibliography}{10}

\bibitem{albrecht_autonomous_2018}
Stefano~V. Albrecht and Peter Stone.
\newblock Autonomous agents modelling other agents: {A} comprehensive survey
  and open problems.
\newblock {\em Artificial Intelligence}, 258:66--95, May 2018.

\bibitem{heard_diagnostic_2018}
Jamison Heard, Rachel Heald, Caroline~E. Harriott, and Julie~A. Adams.
\newblock A {Diagnostic} {Human} {Workload} {Assessment} {Algorithm} for
  {Human}-{Robot} {Teams}.
\newblock In {\em Companion of the 2018 {ACM}/{IEEE} {International}
  {Conference} on {Human}-{Robot} {Interaction}}, pages 123--124, Chicago IL
  USA, March 2018. ACM.

\bibitem{zhao_diverse_2020}
He~Zhao and Richard~P. Wildes.
\newblock On {Diverse} {Asynchronous} {Activity} {Anticipation}.
\newblock In Andrea Vedaldi, Horst Bischof, Thomas Brox, and Jan-Michael Frahm,
  editors, {\em Computer {Vision} – {ECCV} 2020}, Lecture {Notes} in
  {Computer} {Science}, pages 781--799, Cham, 2020. Springer International
  Publishing.

\bibitem{ng_forecasting_2020}
Y.~B. Ng and B.~Fernando.
\newblock Forecasting {Future} {Action} {Sequences} {With} {Attention}: {A}
  {New} {Approach} to {Weakly} {Supervised} {Action} {Forecasting}.
\newblock {\em IEEE Transactions on Image Processing}, 29:8880--8891, 2020.
\newblock Conference Name: IEEE Transactions on Image Processing.

\bibitem{abu_farha_uncertainty-aware_2019}
Yazan Abu~Farha and Juergen Gall.
\newblock Uncertainty-{Aware} {Anticipation} of {Activities}.
\newblock In {\em 2019 {IEEE}/{CVF} {International} {Conference} on {Computer}
  {Vision} {Workshop} ({ICCVW})}, pages 1197--1204, Seoul, Korea (South),
  October 2019. IEEE.

\bibitem{morais_learning_2020}
Romero Morais, Vuong Le, Truyen Tran, and Svetha Venkatesh.
\newblock Learning to {Abstract} and {Predict} {Human} {Actions}.
\newblock {\em arXiv:2008.09234 [cs]}, August 2020.
\newblock arXiv: 2008.09234.

\bibitem{suris_learning_2021}
Didac Suris, Ruoshi Liu, and Carl Vondrick.
\newblock Learning the {Predictability} of the {Future}.
\newblock pages 12607--12617, 2021.

\bibitem{stein_combining_2013}
Sebastian Stein and Stephen~J. McKenna.
\newblock Combining embedded accelerometers with computer vision for
  recognizing food preparation activities.
\newblock In {\em Proceedings of the 2013 {ACM} international joint conference
  on {Pervasive} and ubiquitous computing}, {UbiComp} '13, pages 729--738, New
  York, NY, USA, September 2013. Association for Computing Machinery.

\bibitem{Farha2018b}
Yazan~Abu Farha, Alexander Richard, and Juergen Gall.
\newblock When will you do what? - {Anticipating} {Temporal} {Occurrences} of
  {Activities}.
\newblock In {\em Proceedings of the {IEEE} {Computer} {Society} {Conference}
  on {Computer} {Vision} and {Pattern} {Recognition}}, pages 5343--5352, 2018.
\newblock ISSN: 10636919.

\bibitem{sener_temporal_2020}
Fadime Sener, Dipika Singhania, and Angela Yao.
\newblock Temporal {Aggregate} {Representations} for {Long}-{Range} {Video}
  {Understanding}.
\newblock {\em arXiv:2006.00830 [cs]}, July 2020.
\newblock arXiv: 2006.00830.

\bibitem{piergiovanni_adversarial_2020}
A.~J. Piergiovanni, Anelia Angelova, Alexander Toshev, and Michael~S. Ryoo.
\newblock Adversarial {Generative} {Grammars} for {Human} {Activity}
  {Prediction}.
\newblock In Andrea Vedaldi, Horst Bischof, Thomas Brox, and Jan-Michael Frahm,
  editors, {\em Computer {Vision} – {ECCV} 2020}, Lecture {Notes} in
  {Computer} {Science}, pages 507--523, Cham, 2020. Springer International
  Publishing.

\bibitem{sener_zero-shot_2019}
Fadime Sener and Angela Yao.
\newblock Zero-{Shot} {Anticipation} for {Instructional} {Activities}.
\newblock In {\em 2019 {IEEE}/{CVF} {International} {Conference} on {Computer}
  {Vision} ({ICCV})}, pages 862--871, Seoul, Korea (South), October 2019. IEEE.

\bibitem{guan_generative_2020}
Jiaqi Guan, Ye~Yuan, Kris~M. Kitani, and Nicholas Rhinehart.
\newblock Generative {Hybrid} {Representations} for {Activity} {Forecasting}
  {With} {No}-{Regret} {Learning}.
\newblock In {\em 2020 {IEEE}/{CVF} {Conference} on {Computer} {Vision} and
  {Pattern} {Recognition} ({CVPR})}, pages 170--179, Seattle, WA, USA, June
  2020. IEEE.

\bibitem{gammulle_forecasting_2019}
Harshala Gammulle, Simon Denman, Sridha Sridharan, and Clinton Fookes.
\newblock Forecasting {Future} {Action} {Sequences} with {Neural} {Memory}
  {Networks}.
\newblock {\em arXiv:1909.09278 [cs]}, September 2019.
\newblock arXiv: 1909.09278.

\bibitem{gal_dropout_2016}
Yarin Gal and Zoubin Ghahramani.
\newblock Dropout as a {Bayesian} {Approximation}: {Representing} {Model}
  {Uncertainty} in {Deep} {Learning}.
\newblock In {\em Proceedings of {The} 33rd {International} {Conference} on
  {Machine} {Learning}}, pages 1050--1059. PMLR, June 2016.
\newblock ISSN: 1938-7228.

\bibitem{mehrasa_variational_2019}
Nazanin Mehrasa, Akash~Abdu Jyothi, Thibaut Durand, Jiawei He, Leonid Sigal,
  and Greg Mori.
\newblock A {Variational} {Auto}-{Encoder} {Model} for {Stochastic} {Point}
  {Processes}.
\newblock In {\em 2019 {IEEE}/{CVF} {Conference} on {Computer} {Vision} and
  {Pattern} {Recognition} ({CVPR})}, pages 3160--3169, Long Beach, CA, USA,
  June 2019. IEEE.

\bibitem{Farha2019b}
Yazan~Abu Farha and Juergen Gall.
\newblock {MS}-{TCN}: {Multi}-{Stage} {Temporal} {Convolutional} {Network} for
  {Action} {Segmentation}.
\newblock 2019.
\newblock arXiv: 1903.01945.

\bibitem{lakshminarayanan_simple_2017}
Balaji Lakshminarayanan, Alexander Pritzel, and Charles Blundell.
\newblock Simple and {Scalable} {Predictive} {Uncertainty} {Estimation} using
  {Deep} {Ensembles}.
\newblock In {\em Advances in {Neural} {Information} {Processing} {Systems}},
  volume~30. Curran Associates, Inc., 2017.

\bibitem{gast_lightweight_2018}
Jochen Gast and Stefan Roth.
\newblock Lightweight {Probabilistic} {Deep} {Networks}.
\newblock In {\em 2018 {IEEE}/{CVF} {Conference} on {Computer} {Vision} and
  {Pattern} {Recognition}}, pages 3369--3378, Salt Lake City, UT, June 2018.
  IEEE.

\bibitem{li_ms-tcn_2020}
Shi-Jie Li, Yazan AbuFarha, Yun Liu, Ming-Ming Cheng, and Juergen Gall.
\newblock {MS}-{TCN}++: {Multi}-{Stage} {Temporal} {Convolutional} {Network}
  for {Action} {Segmentation}.
\newblock {\em IEEE Transactions on Pattern Analysis and Machine Intelligence},
  pages 1--1, 2020.
\newblock Conference Name: IEEE Transactions on Pattern Analysis and Machine
  Intelligence.

\bibitem{Chang2019}
Chien~Yi Chang, De~An Huang, Yanan Sui, Li~Fei-Fei, and Juan~Carlos Niebles.
\newblock {D3TW}: {Discriminative} differentiable dynamic time warping for
  weakly supervised action alignment and segmentation.
\newblock {\em Proceedings of the IEEE Computer Society Conference on Computer
  Vision and Pattern Recognition}, 2019-June:3541--3550, 2019.
\newblock arXiv: 1901.02598 ISBN: 9781728132938.

\bibitem{Kuehne2014a}
Hilde Kuehne, Ali Arslan, and Thomas Serre.
\newblock The language of actions: {Recovering} the syntax and semantics of
  goal-directed human activities.
\newblock {\em Proceedings of the IEEE Computer Society Conference on Computer
  Vision and Pattern Recognition}, pages 780--787, 2014.
\newblock ISBN: 9781479951178.

\bibitem{long_searching_2020}
Teng Long, Pascal Mettes, Heng~Tao Shen, and Cees G.~M. Snoek.
\newblock Searching for {Actions} on the {Hyperbole}.
\newblock In {\em 2020 {IEEE}/{CVF} {Conference} on {Computer} {Vision} and
  {Pattern} {Recognition} ({CVPR})}, pages 1138--1147, Seattle, WA, USA, June
  2020. IEEE.

\bibitem{wu_hierarchical_2019}
Cinna Wu, Mark Tygert, and Yann LeCun.
\newblock A hierarchical loss and its problems when classifying
  non-hierarchically.
\newblock {\em PloS One}, 14(12):e0226222, 2019.

\bibitem{farha_long-term_2020}
Yazan~Abu Farha, Qiuhong Ke, Bernt Schiele, and Juergen Gall.
\newblock Long-{Term} {Anticipation} of {Activities} with {Cycle}
  {Consistency}.
\newblock {\em arXiv:2009.01142 [cs]}, September 2020.
\newblock arXiv: 2009.01142.

\bibitem{ke_time-conditioned_2019}
Qiuhong Ke, Mario Fritz, and Bernt Schiele.
\newblock Time-{Conditioned} {Action} {Anticipation} in {One} {Shot}.
\newblock In {\em 2019 {IEEE}/{CVF} {Conference} on {Computer} {Vision} and
  {Pattern} {Recognition} ({CVPR})}, pages 9917--9926, Long Beach, CA, USA,
  June 2019. IEEE.

\bibitem{loh_long-term_2022}
Siyuan~Brandon Loh, Debaditya Roy, and Basura Fernando.
\newblock Long-{Term} {Action} {Forecasting} {Using} {Multi}-{Headed}
  {Attention}-{Based} {Variational} {Recurrent} {Neural} {Networks}.
\newblock pages 2419--2427, 2022.

\end{thebibliography}
\end{document}